\pdfoutput=1

\documentclass[11pt]{article}

\usepackage{naacl2021}

\usepackage{times}
\usepackage{latexsym}

\usepackage[T1]{fontenc}

\usepackage[utf8]{inputenc}

\usepackage{microtype}

\usepackage{amsfonts}
\usepackage{booktabs}
\usepackage{xcolor}

\usepackage{url}
\usepackage{latexsym}
\usepackage{bm}
\usepackage{algorithm}
\usepackage{algpseudocode}
\usepackage{graphicx}
\usepackage{amsmath}
\usepackage{amssymb}
\usepackage{caption}
\usepackage{color}
\usepackage{threeparttable}
\usepackage{microtype}
\usepackage{multirow}
\usepackage{color}
\usepackage[switch]{lineno}

\title{Constrained Text Generation with Global Guidance -- Case Study on CommonGen}

\author{Yixian Lu$^1$\footnotemark[1], Liwen Zhang$^3$\footnotemark[1], Wenjuan Han$^4$, Yue Zhang$^2$, Kewei Tu$^3$ \\
  $^1$Interactive Entertainment Group, Tencent Inc. \, $^2$School of Engineering, Westlake University \\
  $^3$School of Information Science and Technology, ShanghaiTech University \\
  $^4$School of Computing, National University of Singapore \\
  \texttt{easingliu@tencent.com},
  \texttt{\{zhanglw1,tukw\}@shanghaitech.edu.cn} \\
  \texttt{hanwj0309@gmail.com}, 
  \texttt{yue.zhang@wias.org.cn}}

\begin{document}
\maketitle

\renewcommand{\thefootnote}{\fnsymbol{footnote}}
\footnotetext[1]{Equal contributions.}

\begin{abstract}
This paper studies constrained text generation, which is to generate sentences under certain pre-conditions. 
We focus on CommonGen, the task of generating text based on a set of concepts, as a representative task of constrained text generation. 
Traditional methods mainly rely on supervised training to maximize the likelihood of target sentences.
However, global constraints such as common sense and coverage cannot be incorporated into the likelihood objective of the autoregressive decoding process. 
In this paper, we consider using reinforcement learning to address the limitation, measuring global constraints including fluency, common sense and concept coverage with a comprehensive score, which serves as the reward for reinforcement learning. 
Besides, we design a guided decoding method at the word, fragment and sentence levels. 
Experiments demonstrate that our method significantly increases the concept coverage and outperforms existing models in various automatic evaluations.
\end{abstract}

\section{Introduction}

With the rise of deep learning methods, the task of natural language generation has received a surge of research interests \cite{DBLP:journals/jair/GattK18,reiter2000building}. 
Sequence to sequence models have been used for meaning-to-text generation \cite{song-etal-2020-structural,DBLP:conf/acl/GildeaWZS18,DBLP:conf/naacl/DamonteC19}, data-to-text generation \cite{DBLP:conf/aaai/Puduppully0L19} and text-to-text generation tasks such as abstractive summarization \cite{liu-lapata-2019-text, tan2017abstractive} and machine translation \cite{DBLP:conf/nips/SutskeverVL14,bahdanau2014neural}. The typical model is trained using a given set of output texts, using a cross-entropy loss to maximize the probability of each token given its preceding text. 
In other words, models are trained to mimic a ``standard'' human-written token sequence.

While such methods have been shown effective, they intuitively can be limited by forcing one correct answer on the model for each input. In contrast, the nature of languages suggests that there can be a range of different ways of expressing the same meaning. As a result, one solution can be to allow a model to generate outputs in freedom, giving feedback on the output quality. The recent investigation of human rewards on text summarizers \citep{stiennon2020learning} shows that such methods can largely outperform traditional methods.

While human feedback can be highly costly to obtain, we consider automatic ways for collecting feedback. In particular, we focus on constrained text generation, which constructs sentences under certain pre-conditions such as content \cite{DBLP:journals/corr/abs-1911-03705}, style \cite{DBLP:conf/nips/ShenLBJ17} and topics \cite{DBLP:conf/ijcai/FengLL0SL18,DBLP:journals/corr/abs-1903-07137}. 
In the broadest sense, most text generation tasks are constrained tasks. For example, the constraints for machine translation includes fluency and adequacy. Therefore our investigation can be generalized.

We focus on a commonsense text generation task \cite{DBLP:journals/corr/abs-1911-03705}, which asks the model to generate a natural text description given a set of common concepts. 
For example, if the input concept set is \{\textit{kid}, \textit{room}, \textit{dance}\}, a plausible output sentence is ``\textit{The kid loves to dance in her own room}.''. To accomplish such a task, a generation model should understand the correlation between the given concepts and insert necessary words to produce a fluent sentence. As there can be different ways to interpret the correlation between the input concepts, the corresponding output sentence is not unique in terms of both meaning and phrasing. Thus the task provides a convenient playground for investigating reward-based training.

There are three main constraints for the task. First, the output should be fluent and grammatical. Second, the sentence should follow common sense. Third, The output should contain as many input concepts as possible. Correspondingly, we consider three automatic measures of quality, namely fluency, common sense and concept coverage. For fluency, we measure the perplexity of output sentences using GPT-2 \cite{DBLP:journals/corr/abs-2005-09123} assuming that fluent sentences have low perplexity. For measuring common sense, we fine-tune GPT-2 with \texttt{ConceptNet} \cite{DBLP:conf/aaai/SpeerCH17}, by following \citet{DBLP:conf/acl/WangLZLG19}. Coverage is estimated by simple string matching.

The three types of automatic rewards are used to tune a UniLM \cite{DBLP:conf/nips/00040WWLWGZH19,DBLP:journals/corr/abs-2002-12804} model, which is a sequence to sequence method built on Transformer \cite{DBLP:conf/nips/VaswaniSPUJGKP17}. In particular, we first train a model by following the traditional method, maximizing the likelihood of a given set of human-written outputs. After obtaining the traditional model to its optimal performance, we use reinforcement learning \cite{Williams1992SimpleSG} to further tune the parameters using the above rewards. This allows the system to freely generate outputs, receive feedback, and correct itself accordingly.

In addition, we design a guided decoding method at three levels: word, fragment, and sentence. 
At the word level, the generative model and the GPT-2 work together to predict the next word. At the fragment level, we construct a extra beam to save sentence fragments with the highest scores. At the sentence level, the score is used to re-rank the sentences produced by beam search.


Our experiments demonstrate that the guided training and decoding processes effectively improve the generation performance. Our method significantly increases the concept coverage and outperforms existing models in various automatic evaluations. To our knowledge, we are the first to construct a CommonGen method based on global constraints. We will release our code and model later.  


\section{Related Work}
\citet{DBLP:conf/acl/WangLZLG19} use ELMo \cite{DBLP:conf/naacl/PetersNIGCLZ18} and BERT\cite{DBLP:conf/naacl/DevlinCLT19} to calculate the normalized likelihood of a sentence as the measurement of common sense, and ConceptNet is used for fine-tuning. 
We consider a similar fine-tuning method as proposed in \citet{DBLP:conf/acl/WangLZLG19} but use GPT-2, which we find achieving better performance in common sense judgement.
\citet{DBLP:journals/tacl/GuanHHZZ20} also use ConceptNet to fine-tune GPT-2. They took fine-tuned GPT-2 as the generative model, and fine-tuning of GPT-2 on ConceptNet is to improve the long-term relevance among the sentences in generated stories. In contrast, the fine-tuned GPT-2 works as a discriminator in our method to measure the common sense of a single sentence.


Previous work has also considered enforcing global constraints with reinforcement learning for text generation. \citet{DBLP:conf/emnlp/WuTQLL18} and  \citet{DBLP:conf/emnlp/DuJ19} consider automatic evaluations such as BLEU and METEOR. \citet{DBLP:conf/aaai/YuZWY17} consider the naturality of the sentences measured by an adversarial discriminator. 
In contrast to their work, we make use of a set of different metrics to comprehensively evaluate the quality of output sentences, without the need of a reference output.
\citet{DBLP:journals/corr/abs-1909-08593} consider the human evaluation of the generated sentence. 
We differ from the previous work in that we use a range of automatic measures to score the fluency and common sense in guided reinforcement learning.

\section{Task and Model}

The input of CommonGen is a set of $M$ concepts $\mathbf{X} = \{x_1,...,x_M\}$, where every $x_i$ indicates a noun or verb. The target output is a natural sentence containing these concepts $\mathbf{Y} = [y_1, ...,y_N]$, where $y_i$ is a word. We expect that the output sentence is grammatical, fluent and does not violate common sense. The task is to learn a function to map the concept set $\textbf{X}$ to a proper sentence $\textbf{Y}$.

\citet{DBLP:journals/corr/abs-1911-03705} solve this task supervised learning. Given a training sample $[\mathbf{X}, \mathbf{Y}]$, the objective is
\begin{flalign*}
\ell_{mle} &= -\log \text{P}(\mathbf{Y}|\mathbf{X}) \\
         &= -\sum_{t=1}^{N} \log \text{P}(y_t|y_{<t}, \mathbf{X}).
\end{flalign*}

In this paper, our generative model is UniLM \cite{DBLP:conf/nips/00040WWLWGZH19}.

\section{Global Guidance}\label{sec:reward}


\subsection{Fluency and Common Sense}



Inspired by \citet{holtzman2018learning,han-etal-2020-adversarial}, we use perplexity on GPT-2 \citep{radford2019language} to evaluate the fluency and common sense of the generated sentences.

Since the domain of the perplexity is $(0, \inf)$, perplexity is not a suitable value as reward.
we use a linear function to normalize it to $ [0,1]$, 
\begin{align*}
    S_{\text{PPL}}(\mathbf{Y})& = 
    \begin{cases}
        0 ,&  \text{PPL}(\mathbf{Y}) \leq \text{L}\\
        \frac{\text{U}-\text{PPL}(\mathbf{Y}) }{\text{U}-\text{L}} ,& \text{U} < \text{PPL}(\mathbf{Y}) < \text{L}\\
        1 ,&  \text{PPL}(\mathbf{Y}) \geq \text{U}
    \end{cases}
\end{align*}
where``U'' and ``L'' are the pre-defined upper and lower bounds, respectively.



As \citet{DBLP:conf/acl/WangLZLG19} shows, the performance of pre-trained language models on common sense can be improved, if the model is fine-tuned on \texttt{ConceptNet} \cite{DBLP:conf/aaai/SpeerCH17}.
Thus,we fine-tune GPT-2 on \texttt{ConceptNet} and evaluate the ability of common sense judgement with task 4 \cite{wang-etal-2020-semeval} in SemEval-2020. 
The accuracy of GPT-2 improves from 76.23\% to 83.35\%. 
The reward of fine-tuned GPT-2 is $S_{\text{PPL}, \text{F}}(\mathbf{Y})$.

\subsection{Concept Coverage}

For CommonGen, the generated sentence is encouraged to contain as many input concepts as possible, 
so we also consider the coverage of the input concepts as a global constraint.
%
To calculate the coverage, we match the lemmas of the output words and the input concepts.\footnote{We apply Pattern \cite{DBLP:journals/jmlr/SmedtD12} to obtain the lemma of a word.} 
The definition of concept coverage is
\begin{flalign*}
S_{\text{cov}}(\mathbf{X},\mathbf{Y}) = \frac{|\text{Captured concepts}|}{|\text{Input concepts}|} \in [0, 1]
\end{flalign*}

\subsection{Sentence Length}\label{sec:len_reward}
Intuitively, it is easier to capture more input concepts with more output words. 
In addition, longer sentences tend to get lower GPT-2 perplexity. 

In order to penalize the generated sentences that is two times or more longer than the input, we design the following length score,
\begin{align*}
S_{\text{len}}(\mathbf{X},\mathbf{Y}) = \min\Big(\frac{2 \cdot \text{len}_{\text{input}}}{\text{len}_{\text{output}}}, 1\Big) \in (0, 1]
\end{align*}
where $\text{len}_{\text{input}}$ and $\text{len}_{\text{output}}$ indicate the number of input concepts and the length of the output sentence, respectively. 

\subsection{Comprehensive Score}
We compute the weighted sum of the above for a comprehensive global score,
\begin{align*}
R(\mathbf{X},\mathbf{Y}) & = w_1 S_{\text{PPL}}(\mathbf{Y}) + w_2 S_{\text{PPL}, \text{F}}(\mathbf{Y}) \\
& + w_3 S_{\text{cov}}(\mathbf{X},\mathbf{Y}) + w_4 S_{\text{len}}(\mathbf{X},\mathbf{Y})  
\end{align*}
where $w_1$, $w_2$, $w_3$, $w_4$ are hyper-parameters. $w_1$ and $w_2$ cannot be non-zero at the same time.

\section{Reinforcement Learning}
We use REINFORCE algorithm \cite{Williams1992SimpleSG}.
The objective function is
\begin{flalign*}
\ell_{rl} &= \text{E}_{\hat{\mathbf{Y}}\sim\text{P}(\hat{\mathbf{Y}}|\mathbf{X})} R(\hat{\mathbf{Y}}, \mathbf{X})
\end{flalign*}
where $\hat{\mathbf{Y}}$ is a sentence sampled based on $\mathbf{X}$ by the generative model, and $R$ is reward.

The gradient can be estimated using Monte-Carlo sampling. At each step of the algorithm, we sample multiple $\hat{\mathbf{Y}}$ based on the generative model, calculate the reward $R(\hat{\mathbf{Y}}, \mathbf{X})$ and compute the gradient. Specifically, for each sampled pair $(\hat{\mathbf{Y}}, \mathbf{X})$, we calculate the gradient as 
\begin{flalign*}
\nabla_\theta \ell_{rl}(\hat{\mathbf{Y}}, \mathbf{X}) &= \big(R(\hat{\mathbf{Y}}, \mathbf{X })-\overline{R}\big)\nabla_\theta \log P(\hat{\mathbf{Y}}|\mathbf{X})
\end{flalign*}
where $\theta$ is the parameter to be optimized in the generative model. In order to reduce the variance of the gradient, we subtract a baseline value $\overline{R}$ from the reward. $\overline{R}$ is the estimated average reward over multiple sampled sentences at every step as described by \citet{DBLP:conf/emnlp/WuTQLL18}.

Inspired by the \texttt{Dagger} algorithm \cite{DBLP:journals/jmlr/RossGB11}, which performs better in tasks such as paraphrase generation. 

By increasing the probabilities of proper sentences \cite{DBLP:conf/emnlp/DuJ19},
we sample multiple high quality sentences with beam search and calculate $R(\mathbf{X}, \mathbf{Y})$ based on these sentences, which is similar to the $\epsilon$-greedy algorithm \cite{DBLP:conf/ki/TokicP11}. 

\section{Decoding}
The global guidance can also be used to improve the decoding process.
We construct the guided decoding processes at three levels. 

\subsection{Word level: Interpolation} 
We use the interpolation of the distribution of generation model $P_{gm}(y_t|y_{<t}, \mathbf{X})$ and GPT-2 $P_{gpt}(y_t|y_{<t})$ to predict the next word:
\begin{align}
P(y_t|y_{<t}, \mathbf{X}) & =  \alpha P_{gm}(y_t|y_{<t}, \mathbf{X}) \nonumber \\
                          & + (1-\alpha) P_{gpt}(y_t|y_{<t}) \label{eq:m_dist}
\end{align}
where $\alpha$ is hyper-parameter.


\subsection{Fragment level: Guided Beam Search} 
Beam search \cite{DBLP:conf/aclnmt/FreitagA17} generates a sentence with the (locally) maximum likelihood. 
For a beam with width $K$, in each step of generation, $K^2$ candidate sentence fragments are sampled, and the $K$ fragments with the highest likelihood are saved. 
Guided beam search is a similar sampling strategy but using the comprehensive score instead of the likelihood.
In every step of guided beam search, we compute the comprehensive scores of the fragment candidates and save the $K$ fragments with the highest scores. 

Specifically, we construct a guided beam $B_g$ that works in parallel with the traditional beam $B$. $B_g$ is used to save the sentence fragments with the highest comprehensive scores $R(\mathbf{X},\mathbf{Y})$ in each decoding step, which is shown in the algorithm \ref{al:guided_beam}. 
As the perplexity cannot measure fragments properly, 
the comprehensive score $R(\mathbf{X},\mathbf{Y})$ only contains $S_{\text{cov}}(\mathbf{X},\mathbf{Y})$ and $S_{\text{len}}(\mathbf{X},\mathbf{Y})$. 
Generally, the sentence fragments capturing more concepts will be saved in the guided beam. When multiple sentences have the same concept coverage, we keep the shorter fragments.


\subsection{Sentence Level: Re-ranking} 
We use the comprehensive score $R$ to re-rank the guided beam search results, and select the sentence with the highest score.
Intuitively, the sentence with the highest score meets the constraint of CommonGen better than the sentence with the largest likelihood.
Note that, the weights $w_1 \cdots w_4$ of the comprehensive score during re-ranking is different of the weights during training.  

\begin{algorithm}[t]
\caption{Guided Beam Search}
\label{al:guided_beam}
\begin{algorithmic}[1]
\Require
Input concept set $\mathbf{X}$, the generative model $G$, beam size $K$, the maximum length of generated sentence $MaxStep$.
\State Initialize $B = \{\}$, $B_g = \{\}$.
\State Predict the first word by the generative model $G$ according to $\mathbf{X}$, and gets $K$ words with the highest likelihood.
\State Update $B$ to save the $K$ words obtained in the previous step, and set $B_g = B$.
\For {$t=2,...,MaxStep$}
\State $B$ and $B_g$ provide a total of $2K$ generated sentence fragments. 
\State Predict the next word  by $G$ according to these fragments.
\State For each fragment, select $K$ predicted words with the highest probabilities, and get a total of $2K^2$ predicted candidate fragments.
\State From the $K^2$ candidate fragments obtained from $B$, select the $K$ fragments with the highest likelihood to update $B$.
\State Consider all $2K^2$ candidate segments and calculate their comprehensive score $R_\text{b}$, then select the $K$ fragments with the largest $R_\text{b}$ to update $B_g$.
\EndFor
\State returns $B$, $B_g$.
\end{algorithmic}
\end{algorithm}

\section{Experiments}
\subsection{Experimental Settings}
\paragraph{Data}
The CommonGen dataset \citep{DBLP:journals/corr/abs-1911-03705} contains different concept sets and each concept set corresponds to 2-5 sentence descriptions.

\begin{table}[t]
\begin{center}
\resizebox{0.45\textwidth}{!}{%
\begin{tabular}{c|c|c|c|c}
\hline
 & $w_1$ & $w_2$ & $w_3$ & $w_4$ \\ \hline
Training & \{0, 20\} & \{0, 20\} & 200 & 0 \\ \hline
\begin{tabular}[c]{@{}c@{}}Guided \\Beam Search\end{tabular} & 0 & 0 & 2000 & 200 \\ \hline
Re-ranking & \{0, 110\} & \{0, 110\} & 210 & 10 \\ \hline
\begin{tabular}[c]{@{}c@{}}Baseline \\ Re-ranking\end{tabular} & 0 & 110 & 110 & 110 \\ \hline
\end{tabular}
}
\end{center}
\caption{Weights for comprehensive score. $w_1$, $w_2$, $w_3$, $w_4$ are weighs for $S_{\text{PPL}}$, $S_{\text{PPL, F}}$, $S_{\text{cov}}$ and $S_{\text{len}}$, respectively. $\{0, 100\}$ in table means that $w_1$ and $w_2$ cannot be non-zero at same time. Baseline Re-rank means the weights used for guided decoding baseline model.}
\label{tab:rl_weights}
\end{table}

\begin{table*}[ht]
\begin{center}
\resizebox{1.0\textwidth}{!}{
\begin{tabular}{l|cc|cc|c|c|c|c|c|c}
 & \multicolumn{2}{c|}{\bf \textsc{ROUGE2/L}} & \multicolumn{2}{c|}{\bf \textsc{BLEU3/4}} & \bf \textsc{METOR} & \bf \textsc{CIDEr} & \bf \textsc{SPICE} & \bf \textsc{Cov} & \bf \textsc{PPL} & \bf \textsc{Len} \\ \hline
UniLM$\ast$ & 21.84 & 42.93 & 39.00 & 28.00 & 29.50 & 14.95 & 30.00 & 88.68 & 79.67 & 12.38 \\
UniLM\tiny{gd} & 22.32 & \textbf{43.70} & 37.60 & 27.20 & 29.80 & 15.08 & 30.70 & 94.38 & 68.22 & 11.36 \\ \hline
RLs+GPT2F & 21.46 & 42.80 & 37.40 & 26.80 & 28.90 & 14.42 & 29.70 & 87.99 & 86.87 & 11.55 \\
RLs+GPT2F\tiny{gd} & 21.86 & 42.00 & 35.70 & 25.70 & 30.70 & 14.07 & 31.30 & 96.07 & 45.14 & 14.70 \\ \hline
RLb+GPT2 & 21.99 & 43.25 & 38.50 & 27.60 & 29.60 & 15.12 & 30.60 & 90.26 & 81.94 & 11.93 \\
RLb+GPT2\tiny{gd} & \underline{22.62} & \underline{43.56} & \underline{39.20} & 28.30 & \underline{31.00} & \underline{15.89} & \underline{31.40} & 95.85 & 55.75 & 13.31 \\
RLb+GPT2F & 21.84 & 43.27 & 38.70 & 27.70 & 29.70 & 15.21 & 30.80 & 90.52 & 81.23 & 13.40 \\
RLb+GPT2F\tiny{gd} & \textbf{22.73} & \textbf{43.70} & \textbf{39.70} & \textbf{28.70} & \textbf{31.10} & \textbf{15.94} & \underline{31.40} & \underline{96.73} & 61.37 & 12.95 \\ \hline
UniLM-v2$\dagger$ & 18.24 & 40.62 & 31.30 & 22.10 & 28.10 & 13.10 & 28.10 & 89.13 & - & - \\
BART$\dagger$ & 22.23 & 41.98 & 36.30 & 26.30 & 30.90 & 13.92 & 30.60 & \textbf{97.35} & - & - \\
T5-Large$\dagger$ & 22.01 & 42.97 & 39.00 & \underline{28.60} & 30.10 & 14.96 & \textbf{31.60} & 95.29 & - & - \\ \hline
\end{tabular}
}
\end{center}

\caption{Comparison between the recent models and our models. UniLM$\ast$ is the baseline our implemented. We achieve competitive result than that reported by \citet{DBLP:journals/corr/abs-1911-03705} with UniLM model. Results with $\dagger$ are reported by \citet{DBLP:journals/corr/abs-1911-03705}.
The best models are \textbf{bold} and second best ones are \underline{underlined} within each metric.}\label{tab:Common_result}
\end{table*}

\begin{table*}[t]
\small
\begin{center}
\resizebox{1.0\textwidth}{!}{%
\begin{tabular}{l|cc|cc|c|c|c|c|c|c}
\hline
 & \multicolumn{2}{c|}{\bf \textsc{ROUGE2/L}} & \multicolumn{2}{c|}{\bf \textsc{BLEU3/4}} & \bf \textsc{METOR} & \bf \textsc{CIDEr} & \bf \textsc{SPICE} & \bf \textsc{Cov} & \bf \textsc{PPL} & \bf \textsc{Len} \\ \hline
RLb+GPT2F+Test & 20.68 & 42.45 & 35.10 & 24.20 & 30.60 & 14.77 & \textbf{32.80} & 98.46 & 68.04 & 13.99 \\
RLb+GPT2F$_{\text{gd}}$+Test & 20.94 & 42.33 & 35.30 & 24.80 & 31.30 & 14.66 & \textbf{32.80} & \textbf{98.91} & 47.07 & 14.93 \\ \hline
RLb+GPT2+Test & 20.47 & 42.19 & 34.70 & 23.90 & 30.60 & 14.57 & 32.70 & 98.34 & 72.87 & 14.08 \\
RLb+GPT2$_{\text{gd}}$+Test & 21.06 & 42.43 & 35.80 & 25.30 & 31.40 & 15.07 & \textbf{32.80} & 98.87 & 47.44 & 14.77 \\ \hline
\end{tabular}
}
\end{center}
\caption{Results of model aiming to mitigate domain gap using test inputs. }
\label{tab:train_on_test}
\end{table*}

\paragraph{Hyper-parameters}
We choose the pre-trained Unified Language Model (UniLM, \citet{DBLP:conf/nips/00040WWLWGZH19}) as our generative model.
Following the training strategy of \citet{DBLP:journals/corr/abs-1911-03705}, we first do supervised training based on the ``unilmv1-large-cased'' model\footnote{For the baseline model, we use the code from \url{https:/ /github.com/INK-USC/CommonGen}.}. 
During training, the batch size is 48, the word masking probability is 0.7, the learning rate warm-up is 0.1, and the learning rate is $1e$-$5$.
After 10 epoch of supervised training, we obtain the baseline model.
Based on the baseline model, we then use our comprehensive score to perform reinforcement learning of the baseline for 1 epoch.
All hyper-parameters are the same as in the supervised training, except that the learning rate is set to $1e$-$8$. 
During reinforcement learning and guided decoding, we set the beam search hyper-parameter $K=5$. 
In the word level guided decoding, we set $\alpha=0.3$ in Equation~\ref{eq:m_dist}. 
The weights for the the comprehensive score are shown in Table~\ref{tab:rl_weights}.

\paragraph{Evaluation}
Following \citet{DBLP:journals/corr/abs-1911-03705}, we compare different models using three N-gram based evaluation metrics: BLEU \cite{DBLP:conf/acl/PapineniRWZ02}, ROUGE \cite{lin2004rouge} and METEOR \cite{DBLP:conf/acl/BanerjeeL05}. We additionally adopt CIDEr \cite{DBLP:conf/cvpr/VedantamZP15} and SPICE \cite{DBLP:conf/eccv/AndersonFJG16} which are specially designed for evaluating captioning tasks, as the CommonGen dataset is constructed from video captions. Moreover, we adopt three other metrics associated with the CommonGen task: the average concept coverage rate ``Cov'', the average perplexity calculated by the fine-tuned GPT-2 ``PPL'' and the average sentence length ``Len''. 





\subsection{Main Results}\label{sec:main}

\begin{table*}[t]
\small
\begin{center}
\resizebox{1.0\textwidth}{!}{%
\begin{tabular}{l|cc|cc|c|c|c|c|c|c}
 & \multicolumn{2}{c|}{\bf \textsc{ROUGE2/L}} & \multicolumn{2}{c|}{\bf \textsc{BLEU3/4}} & \bf \textsc{METOR} & \bf \textsc{CIDEr} & \bf \textsc{SPICE} & \bf \textsc{Cov} & \bf \textsc{PPL} & \bf \textsc{Len} \\ \hline
UniLM$\ast$ & 21.84 & 42.93 & 39.00 & 28.00 & 29.50 & 14.95 & 30.00 & 88.68 & 79.67 & 12.38 \\
UniLM\tiny{gd} & 22.32 & 43.70 & 37.60 & 27.20 & 29.80 & 15.08 & 30.70 & 94.38 & 68.22 & 11.36 \\ \hline
Cov & 21.73 & 43.20 & 38.70 & 27.1 & 29.5 & 15.02 & 30.60 & 91.49 & 94.72 & 11.85 \\
Cov\tiny{gd} & 22.67 & 43.16 & \textbf{39.70} & \textbf{28.80} & 31.00 & 15.87 & 31.30 & 96.34 & 60.42 & 12.09 \\ \hline
GPT2F & 21.89 & 43.13 & 38.10 & 27.40 & 29.20 & 14.90 & 30.00 & 88.60 & 80.92 & 11.81 \\
GPT2F\tiny{gd} & 22.37 & 43.20 & 37.40 & 26.90 & 30.90 & 15.12 & 31.10 & 96.01 & \textbf{49.44} & 12.86 \\ \hline
RLb+GPT2F & 21.84 & 43.27 & 38.70 & 27.70 & 29.70 & 15.21 & 30.80 & 90.52 & 81.23 & 13.40 \\
RLb+GPT2F\tiny{gd} & \textbf{22.73} & \textbf{43.70} & \textbf{39.70} & 28.70 & \textbf{31.10} & \textbf{15.94} & \textbf{31.40} & \textbf{96.73} & 61.37 & 12.95 \\ \hline
\end{tabular}
}
\end{center}
\caption{Results using the reward calculated by the perplexity or the coverage rate solely. ``Cov'' means that only the coverage score $S_{\text{cov}}$ is used as the reward in reinforcement learning, ``GPT2F" means that only the perplexity $S_{\text{PPL,F}}$ is used as the reward.}\label{tab:cover_or_ppl}
\end{table*}

\begin{table*}[t]
\small
\begin{center}
\resizebox{1.0\textwidth}{!}{%
\begin{tabular}{l|cc|cc|c|c|c|c|c|c}
\hline
 & \multicolumn{2}{c|}{\bf \textsc{ROUGE2/L}} & \multicolumn{2}{c|}{\bf \textsc{BLEU3/4}} & \bf \textsc{METOR} & \bf \textsc{CIDEr} & \bf \textsc{SPICE} & \bf \textsc{Cov} & \bf \textsc{PPL} & \bf \textsc{Len} \\ \hline
Beam & 21.84 & 43.27 & 38.70 & 27.70 & 29.70 & 15.21 & 30.80 & 90.52 & 81.23 & 12.38 \\ \hline
Beam+R & 22.57 & 43.55 & 39.50 & 28.30 & 30.80 & 15.76 & 31.40 & 94.62 & 58.26 & 13.05 \\ \hline
Beam+M & 21.95 & 43.28 & 39.10 & 28.20 & 29.80 & 15.38 & 30.80 & 90.34 & 74.15 & 12.11 \\ \hline
Beam+M+R & 22.60 & 43.45 & 39.50 & 28.50 & 31.00 & 15.91 & 31.40 & 94.59 & 54.26 & 13.33 \\ \hline
GBeam+M+R & 22.73 & 43.70 & 39.70 & 28.70 & 31.10 & 15.94 & 31.40 & 96.73 & 61.37 & 12.95 \\ \hline
\end{tabular}
}
\end{center}
\caption{Comparison of non-module function in guided decoding method. ``Beam" shows the result of ``RLb+GPT2F", and ``GBeam+M+R" shows the result of ``RLb+GPT2F$_\text{gd}$". ``Beam'' represents the traditional beam search. ``M'' represents the word level guided decoding, i.e. GPT-2 interpolation. ``GBeam'' represents the fragment level guided decoding (i.e. the guided beam search). ``R'' represents the sentence level guided beam search (i.e. re-ranking of the beam search result).}\label{tab:gd}
\end{table*}

We compare our approach with recent state-of-the-art approaches (i.e., ``UniLM$\ast$'', ``UniLM-v2$\dagger$'' \cite{DBLP:journals/corr/abs-2002-12804}, ``BART$\dagger$'' \cite{DBLP:journals/corr/abs-1910-13461} and ``T5$\dagger$'' \cite{DBLP:journals/corr/abs-1910-10683}) and summarize the results in Table \ref{tab:Common_result}. Models listed in Table \ref{tab:Common_result} with the ``RL'' prefix are our models with different architectures. More specifically, ``RLs'' represents that the sentences are randomly sampled by $P(\mathbf{Y}|\mathbf{X})$ in reinforcement learning while ``RLb'' represents that the sentences are sampled using beam search. 
We use the fine-tuned GPT-2 and original GPT-2 to calculate the perplexity score for ``GPT2F'' and ``GPT2'' respectively. 
The model with the subscript ``gd'' denotes that the output are generated by guided decoding while the model without ``gd'' predict a output using a traditional beam search. 

In general, models with guided training and decoding processes effectively improve the generation performance in most automatic evaluation metrics. 
Compared with the major baseline ``UniLM$\ast$", our best-performing model ``RLb+GPT2F$_{\text{gb}}$" improves the scores of ROUGE2 by 0.89, ROUGEL by 0.77, BLEU3 by 0.7, BLEU4 by 0.7, METOR by 1.6, CIDEr by 0.99, SPICE by 1.4 and concept coverage by 8.05. 
It demonstrates the advantage of reward-based tuning using global guidance. 
We also compare our results with three other models on CommonGen reported by \citet{DBLP:journals/corr/abs-1911-03705}: UniLM-v2, BART and T5. 
``RLb+GPT2F$_{\text{gb}}$" achieves the state-of-the-art in CommonGen.

Comparing ``RLs$\cdots$'' and ``RLb$\cdots$'' with the baseline, we find that guided training and decoding with random sampling provides relatively little improvement, while beam search sampling improves the performance by a large margin. This is mainly because beam search sampling can generate higher-quality samples with greater probabilities, improving the efficiency of reinforcement learning.


We also investigate the impact of fine-tuning on GPT-2 by comparing ``RLb+GPT2$\cdots$'' and ``RLb+GPT2F$\cdots$''. It is observed that the models using fine-tuned GPT-2 (``GPT2F'') obtains better performance than their counterparts with original GPT-2 on all the automatic evaluation metrics. This indicates that training with better common sense information improves the generation quality.

\subsection{Results of Using Test Inputs to Mitigate Domain Gap} 
Since reinforcement learning only needs input concepts during training, we continuous train the ``RLb+GPT2F$_{\text{gb}}$'' with the input concepts of the test set to mitigate the domain gap between the training and test datasets. 
In Table \ref{tab:train_on_test}, we report the result of  ``RLb+GPT2F$_{\text{gb}}$'' with a training stage on test inputs. Note that the gold reference sentences of the test set are not used here.
Comparing the result with Table \ref{tab:Common_result}, we can see that training with test inputs achieves a significant increase in concept coverage and SPICE score.


\begin{figure*}[htbp]
\centering
\begin{minipage}[t]{0.32\textwidth}
\centering
\includegraphics[width=5.0cm]{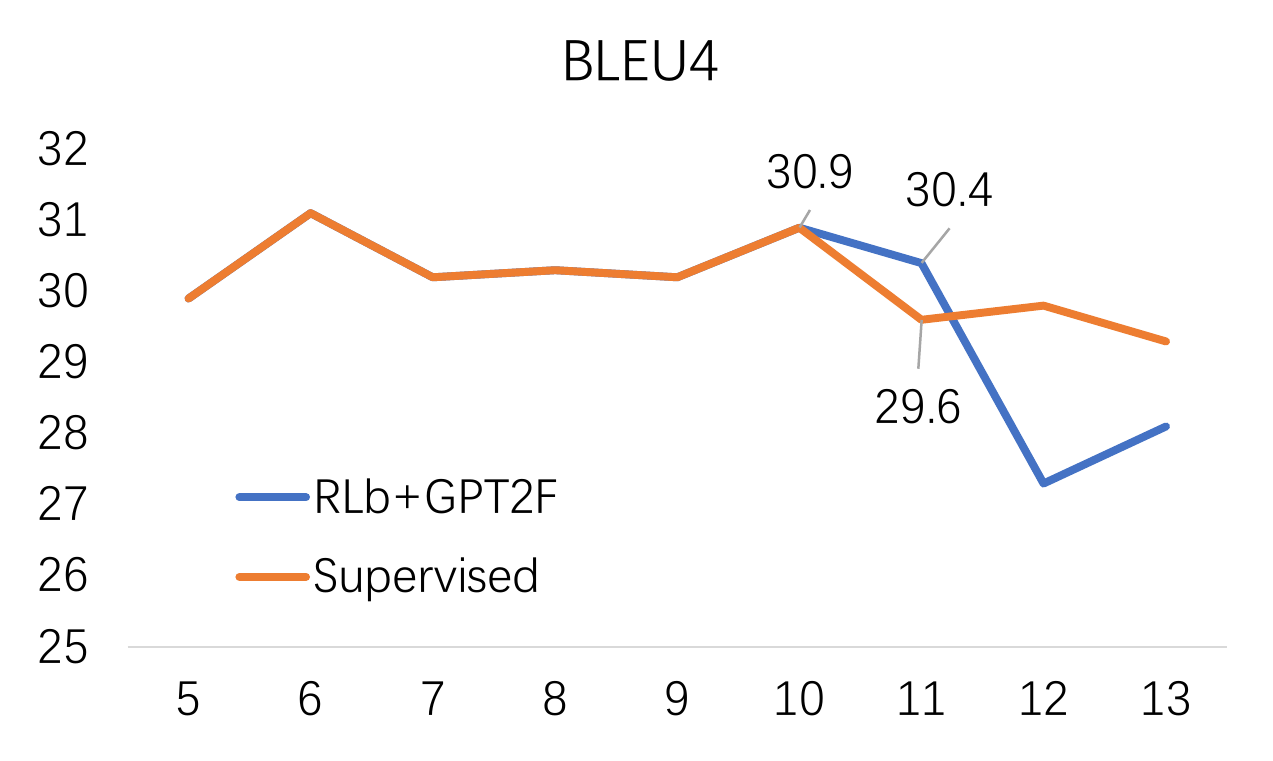}
\captionsetup{labelformat=empty}
\caption{(a)}
\end{minipage}
\begin{minipage}[t]{0.32\textwidth}
\centering
\includegraphics[width=5cm]{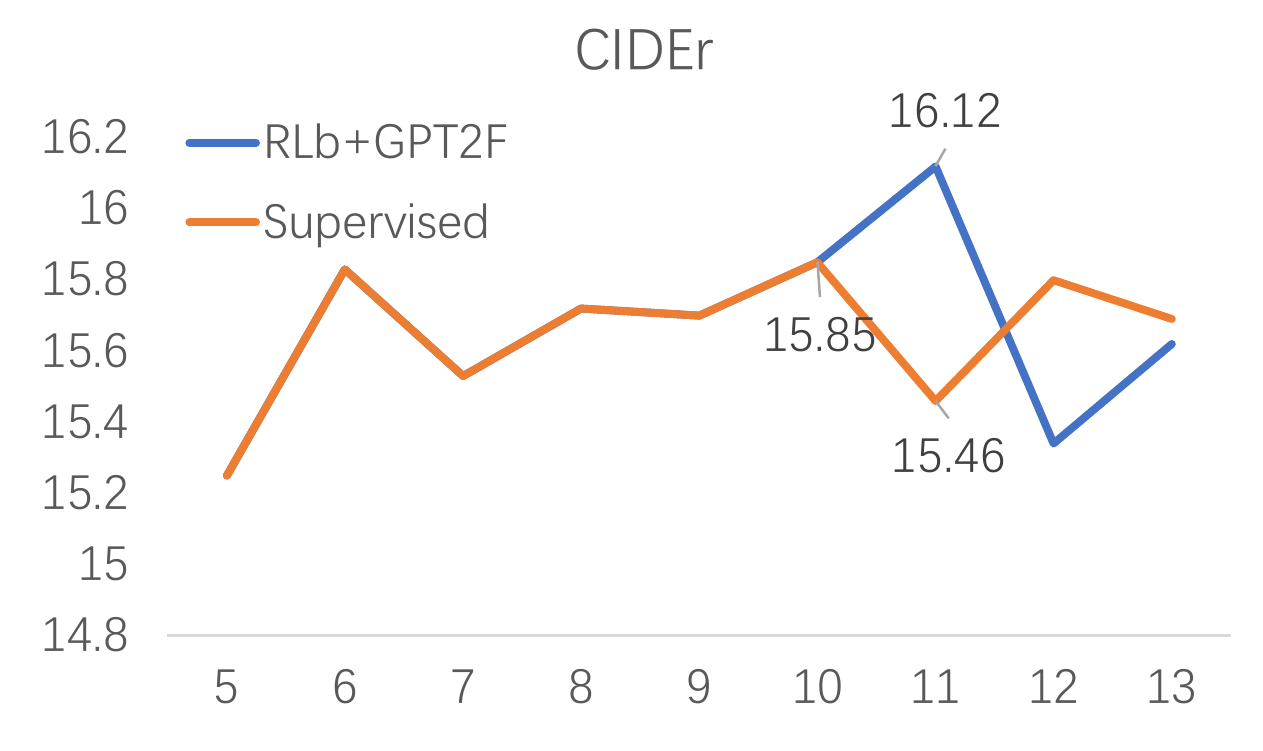}
\captionsetup{labelformat=empty}
\caption{(b)}
\end{minipage}
\begin{minipage}[t]{0.32\textwidth}
\centering
\includegraphics[width=5cm]{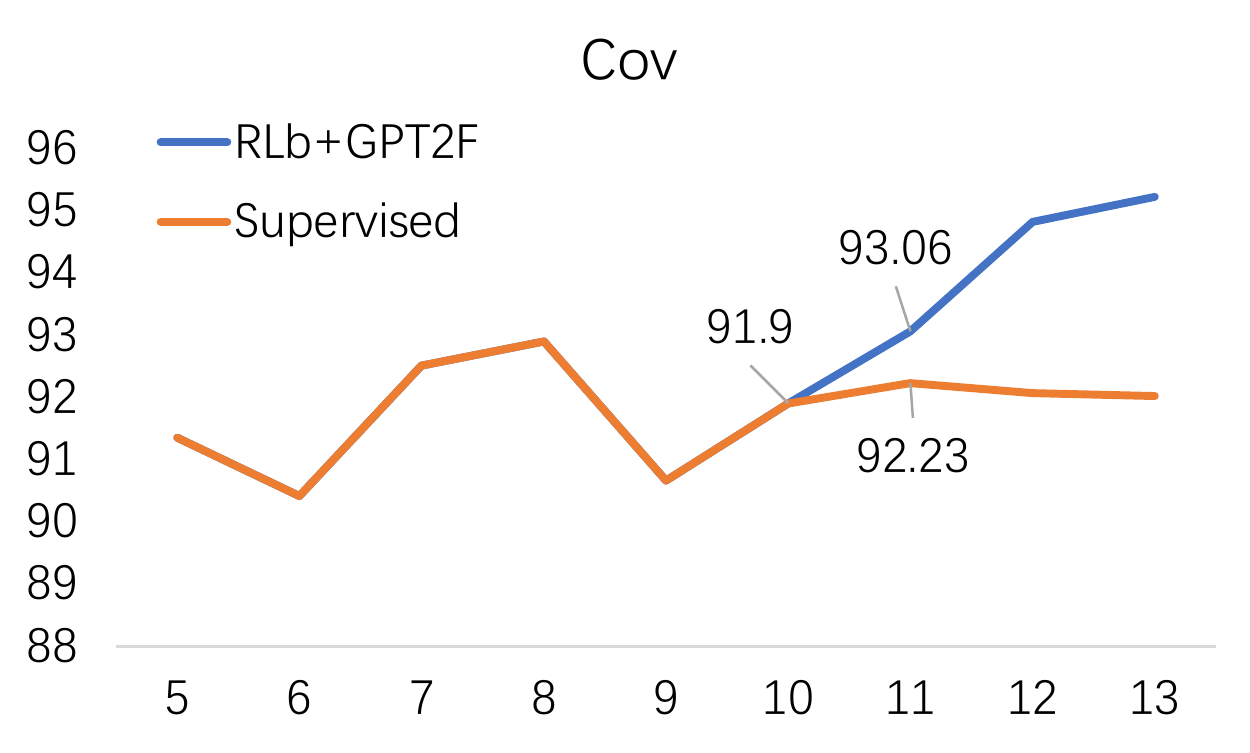}
\captionsetup{labelformat=empty}
\caption{(c)}
\end{minipage}
\caption{Reinforcement learning and supervised learning results on the development dataset. Figure~(a), (b) and (c) show performances on BLEU4, CIDEr and the coverage rate, respectively. }
\label{fig:dev}
\end{figure*}

\begin{figure}[ht]
\resizebox{0.45\textwidth}{!}{%
\centering

\includegraphics{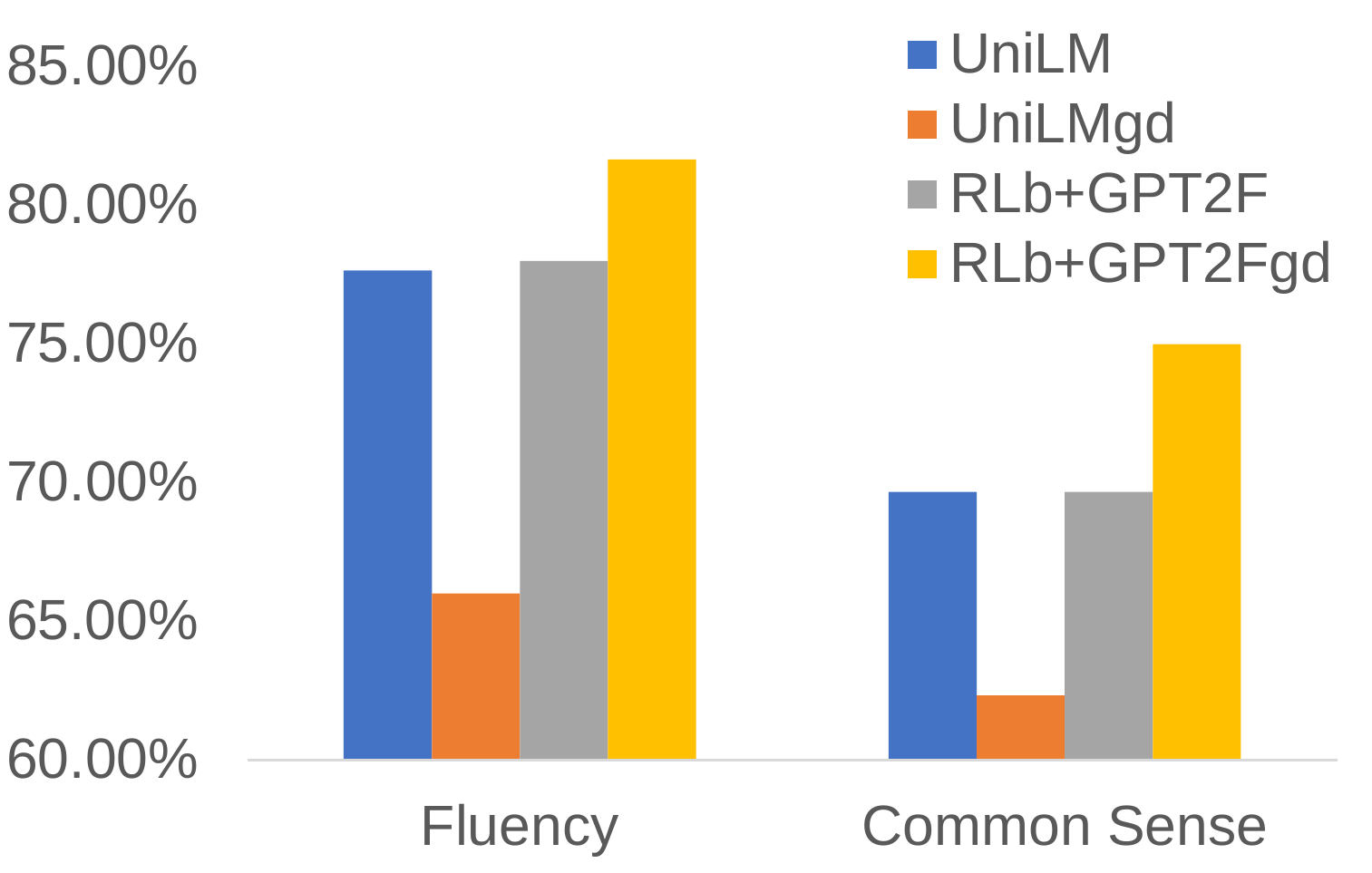}
}
\caption{Human evaluation of fluency and common sense.}
\label{fig:huameval}
\end{figure}
\subsection{{Human Evaluation}}\label{sec:human}
To assess the quality of our method more accurately, we conduct a human evaluation of generation fluency and common sense.
From the examples generated by ``UniLM'', ``UniLM$_\text{gd}$'', ``RLb+GPT2F'' and  ``RLb+GPT2F$_\text{gd}$'', we randomly sample 100 generated sentences. 
In terms of generation fluency and common sense, 3 volunteers are asked to select the best sentences generated by these 4 models. 
We summarize the results in Figure~\ref{fig:huameval}, where each percentage is averaged different volunteers.
Note that we allow ties, so the percentages do not add up to one.

Overall, our ``RLb+GPT2F$_\text{gd}$'' achieves the best results on both metrics.
``UniLM$_\text{gd}$'' achieves the worst results, indicating that direct use of guided decoding on the supervised model may harm performance.
On the other hand, the improvement of ``RLb+GPT2F$_\text{gd}$'' over ``RLb+GPT2F'' shows that using guided decoding is beneficial to our model trained with reinforcement learning.
The discrepancy in the usefulness of guided decoding may be explained by the fact that ``RLb+GPT2F$_\text{gd}$'' is trained using guided training.


\subsection{Ablation}
We study the functions and roles of different components of our method via the ablation experiments. 
\paragraph{Reward in reinforcement learning}
We study the components in the reward function: the perplexity score and coverage score. The results are shown in Table \ref{tab:cover_or_ppl}.  

Comparing the results with the baseline, we can find that, trained with $S_{\text{cov}}$ solely, ``Cov$_\text{gd}$" increases the automatic evaluation and coverage scores from the baseline. 
On the other hand, training with fine-tuned GPT-2 does not improve the automatic evaluation significantly over the baseline, but results in lower perplexity. By comparing ``RLb+GPT2F$_\text{gd}$" with ``Cov$_\text{gd}$", we see that the perplexity score $S_{\text{PPL}}$ further increases the quality of generated sentences, which is reflected 
by the automatic evaluation and perplexity, based on the coverage score, in the reinforcement learning process. 

\paragraph{Guided decoding}
We study the utility of the three levels of guided decoding. The results are shown in Table \ref{tab:gd}.
By comparing ``Beam+M'' and ``Beam'', we conclude that the interpolation of GPT-2 probability improves the overall quality of the generated sentences. It increases the automatic evaluation scores and reduces the perplexity. However, it results in a little drop of the concept coverage, because GPT-2 focus more on the fluency. By comparing ``GBeam+M+R'' with ``Beam+M+R'', 
it is obvious that the guided beam search generally increases the sentence quality and the concept coverage. 
It increases the concept coverage by 2.14 and slightly improves the automatic evaluation and perplexity. 
Guided beam search is significantly more effective than GPT-2 interpolation, implying that the concept coverage plays an important role in global guidance.
By comparing ``Beam+R'' 
and ``Beam+M+R'' with ``Beam'' and ``Beam+M'', we see that the re-ranking process increases the concept coverage by nearly 4. 
It also improves the various automatic evaluations by 0.4$\sim$1. 
Re-ranking makes the most significant improvement compared with guided decoding at the word and fragment levels, because it is based on the global score of the complete sentences.


\subsection{Comparison with Supervised Method}
To investigate how the reinforcement learning method influences the text generation compared with supervised learning, we conduct experiments on the development dataset and summarize the results in Figure~\ref{fig:dev}.
Figure~\ref{fig:dev} (a) (b) and (c) show BLEU4, CIDEr and the coverage rate of different epochs. It shows that training using supervised learning ($>10$ epochs) may hinder the performance. Instead of adhering to supervised training, using reinforcement learning to incorporate global guided information can generally and significantly contribute to the model on various metrics.

\begin{table*}[t]
\small
\begin{center}
\resizebox{1.0\textwidth}{!}{%
\begin{tabular}{lll}
\hline
\bf \textsc{1. Input concept:} & \{stage, perform, routine, music\} &\\\hline
\quad RLb+GPT2F\tiny{gd}: & A group of people \textcolor{blue}{perform} a \textcolor{blue}{routine} on the \textcolor{blue}{stage} during the \textcolor{blue} {music}. & (48.25)\\
\quad UniLM: & A group of people \textcolor{blue}{perform} a \textcolor{blue}{routine} at the \textcolor{blue}{music} festival & (74.17)\\ \hline \hline


\bf \textsc{2. Input concept:} & \{snap, smile, finger, sit\} &\\\hline
\quad RLb+GPT2F\tiny{gd}: & A man \textcolor{blue}{sits} on a bench and \textcolor{blue}{snaps} a \textcolor{blue}{finger} at a \textcolor{blue} {smiling} woman & (45.32)\\
\quad UniLM: & Someone \textcolor{blue}{sits} next to someone and \textcolor{blue}{snaps} a \textcolor{blue}{finger} at him. & (77.18)\\ \hline \hline

\bf \textsc{3. Input concept:} & \{dog, stand, groom, table\}& \\\hline
\quad RLb+GPT2F\tiny{gd}: & A man is \textcolor{blue}{grooming} a \textcolor{blue}{dog} and \textcolor{blue}{standing} next to a \textcolor{blue}{ table}. & (35.16)\\
\quad UniLM: & A man is \textcolor{blue}{grooming} a \textcolor{blue}{dog} and \textcolor{blue}{grooming} it on a \textcolor{blue}{table}. & (85.56) \\ \hline \hline

\bf \textsc{4. Input concept:} & \{throw, ball, pitcher, batter\}& \\\hline
\quad RLb+GPT2F\tiny{gd}: & A \textcolor{blue}{pitcher} \textcolor{blue}{throws} a \textcolor{blue}{ball} to a \textcolor{blue}{batter} & (11.88)\\
\quad UniLM: & A \textcolor{blue}{batter} \textcolor{blue}{throws} a \textcolor{blue}{ball} to the \textcolor{blue}{pitcher}. & (38.54)\\ \hline
\end{tabular}
}
\end{center}
\caption{Case study of an generated sentences from the baseline model ``UniLM'' and ``RLb+GPT2F$_{gd}$''. Value in the bracket is the perplexity of fine-tuned GPT-2.}\label{tab:case_study}
\end{table*}


\begin{figure}[t]
\resizebox{0.45\textwidth}{!}{%
\centering
\includegraphics{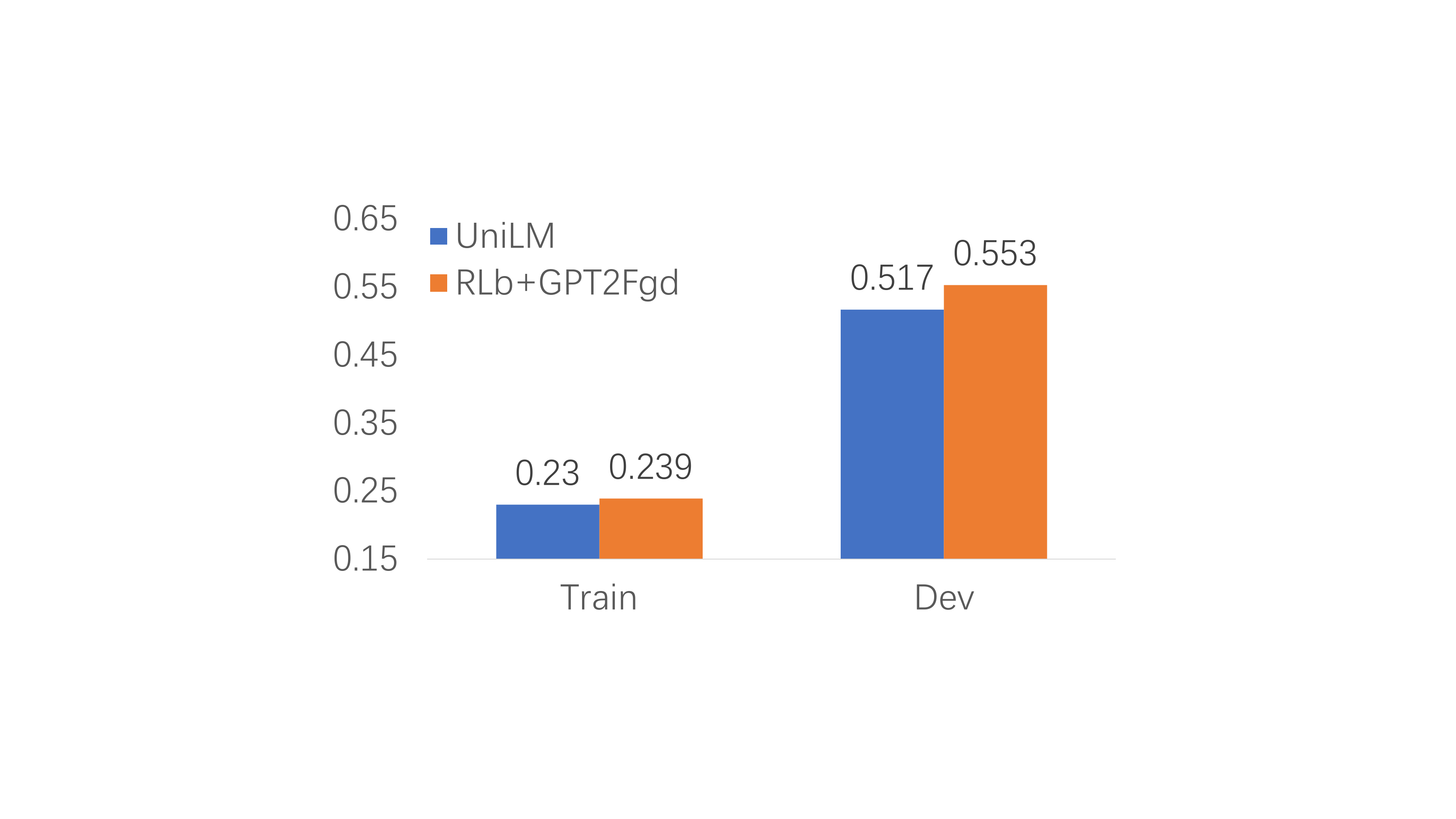}
}
\caption{Minimum edit distance of concepts order between ground truth and generated sentences. Blue bars show the minimum edit distances on training and development dataset between the ground truth and sentences generated by ``UniLM'' . Orange bars show results for ``RLb+GPT2F$_\text{gb}$''.}
\label{fig:ed}
\end{figure}

\subsection{Concept Ordering}

Typical supervised learning using the cross-entropy loss forces the model to mimic a ``standard'' human-written token sequence. The concept order is fixed in the human-written token sequence, which may hinder the diversity of texts generated by the trained model. 
We use reinforcement learning to address this limitation because reinforcement learning provides a reward that is independent of the concept order in the human-written token sequence. 
To empirically verify the advantage of our method in promoting flexibility, we study the concept orders in the sentences generated by the supervised learning based model (``UniLM'') and the reinforcement learning based model (``RLb+GPT2F$_\text{gd}$'').
For each input, we calculate the minimum edit distance between the ordered concept lists drawn respectively from the reference sentence and the generated sentence to measure their difference. 
We compare the average minimum edit distance of ``UniLM'' and ``RLb+GPT2F$_\text{gd}$'' on both the training dataset and the development dataset in Figure~\ref{fig:ed}. The larger minimum edit distance of ``RLb+GPT2F$_\text{gd}$'' in Figure~\ref{fig:ed} suggests that our method provides more freedom in concept ordering and hence may generate more diversified sentences. 

\section{Case Study}\label{sec:case_study}
We provide a case study in Table \ref{tab:case_study}. Given the input concepts, we show the sentence generated by the baseline model ``UniLM'' and our best model ``RLb+GPT2F$_{\text{gd}}$''. For the first three cases, the sentences generated by ``RLb+GPT2F$_{\text{gd}}$'' contain more input concepts than the baseline model, and the coverage rates are $1$. In the first case, our model reasonably included the concept of ``\textit{stage}'' in the sentence 
without undermining the fluency. In the second case, our model not only captures the concept of ``\textit{smile}'', but also solves the problem of ambiguous reference of the sentence by the baseline model. In the third case, compared with the baseline model, our output not only captures ``\textit{stand}'', but also avoids the repetition of ``\textit{grooming}'', which seems to be an improvement in general. In the last case, the sentences generated by the two models are both grammatically correct. But from human's common sense, pitchers are more likely to throw the ball than batters. Therefore, our generated sentences are more in line with common sense. Furthermore, according to the perplexity comparison provided in Table \ref{tab:case_study}, it can be observed that the fine-tuned GPT-2 has the ability to measure the fluency and common sense of the generated sentence.


\section{Conclusion}
We used CommonGen as a case for investigating constrained text generation. A comprehensive score for the sentence measured fluency, common sense, and concept coverage and served as the global guidance of the generative model. We took the comprehensive score as the reward and use reinforcement learning to train the model. We also designed a guided decoding method at word, fragment and sentence levels. Our experiments demonstrated that our method significantly increases the concept coverage compared with the baseline model, and generally improved the scores of various automatic evaluation and human evaluation.


\bibliography{anthology,custom}
\bibliographystyle{acl_natbib}


\end{document}